\documentclass[letterpaper, 10 pt, conference]{ieeeconf}  % Comment this line out if you need a4paper

\IEEEoverridecommandlockouts                            
\usepackage{amsmath}
\usepackage{amsmath}

\usepackage{stfloats} 
\usepackage{afterpage}
\usepackage{booktabs}    % 专业三线表样式
\usepackage{siunitx}     % 数值对齐
\usepackage{makecell}    % 单元格换行
\usepackage{multirow}    % 多行合并
\usepackage{array}       % 增强表格列格式
\usepackage{graphicx}

\usepackage{float}
\usepackage{placeins} % 在你的文档导言（\documentclass之后）添加

\usepackage{etoolbox}
\title{\LARGE \bf
UniDiffGrasp: A Unified Framework Integrating VLM Reasoning and VLM-Guided Part Diffusion for Open-Vocabulary Constrained Grasping with Dual Arms
}

\author{Xueyang Guo\textsuperscript{1*},
Hongwei Hu\textsuperscript{1*},
Chengye Song\textsuperscript{1},
Jiale Chen\textsuperscript{1},
Zilin Zhao\textsuperscript{2},
Yu Fu\textsuperscript{3}, \\ % Optional line break for better layout in PDF
Bowen Guan\textsuperscript{4}, and Zhenze Liu\textsuperscript{1\textdagger}% <-this % stops a space
\thanks{*Equal contribution.}% <-this % stops a space
\thanks{\textsuperscript{\textdagger}Corresponding author.}% <-this % stops a space
\thanks{\textsuperscript{1}School of Communication Engineering, Jilin University, Changchun, China (e-mail: gxy9922@mails.jlu.edu.cn; zzliu@jlu.edu.cn).}%
\thanks{\textsuperscript{2}College of Software Engineering, Jilin University, Changchun, China.}%
\thanks{\textsuperscript{3}School of Mathematics, Jilin University, Changchun, China.}%
\thanks{\textsuperscript{4}College of Electronic Science and Engineering, Jilin University, Changchun, China.}%
}

\begin{document}

\maketitle

\thispagestyle{empty}
\pagestyle{empty}

\newcommand{\comment}[1]{\textbf{\textcolor{red}{#1}}}

%%%%%%%%%%%%%%%%%%%%%%%%%%%%%%%%%%%%%%%%%%%%%%%%%%%%%%%%%%%%%%%%%%%%%%%%%%%%%%%%
\begin{abstract}

Open-vocabulary, task-oriented grasping of specific functional parts, particularly with dual arms, remains a key challenge, as current Vision-Language Models (VLMs), while enhancing task understanding, often struggle with precise grasp generation within defined constraints and effective dual-arm coordination. We innovatively propose UniDiffGrasp, a unified framework integrating VLM reasoning with guided part diffusion to address these limitations. UniDiffGrasp leverages a VLM to interpret user input and identify semantic targets (object, part(s), mode), which are then grounded via open-vocabulary segmentation. Critically, the identified parts directly provide geometric constraints for a Constrained Grasp Diffusion Field (CGDF) using its Part-Guided Diffusion, enabling efficient, high-quality 6-DoF grasps without retraining. For dual-arm tasks, UniDiffGrasp defines distinct target regions, applies part-guided diffusion per arm, and selects stable cooperative grasps. Through extensive real-world deployment, UniDiffGrasp achieves grasp success rates of 0.876 in single-arm and 0.767 in dual-arm scenarios, significantly surpassing existing state-of-the-art methods, demonstrating its capability to enable precise and coordinated open-vocabulary grasping in complex real-world scenarios.

\end{abstract}

%%%%%%%%%%%%%%%%%%%%%%%%%%%%%%%%%%%%%%%%%%%%%%%%%%%%%%%%%%%%%%%%%%%%%%%%%%%%%%%%
\section{INTRODUCTION}

The ambition for robots to seamlessly integrate into human environments as capable assistants hinges on their ability to perform dexterous, task-oriented manipulation. This requires robots not merely to seize an object, but to grasp it by specific functional parts (affordances) conducive to an often implicitly stated goal, such as grasping a screwdriver by its handle for use or a pot by both its handles for stable carrying. This challenge intensifies in open-vocabulary settings, where robots must interpret diverse, sometimes ambiguous, natural language instructions to identify relevant objects and their functional parts, even those unseen during training \cite{3chen2023open,4xu2023joint}. Furthermore, many objects necessitate coordinated dual-arm manipulation due to size, weight, or task requirements, adding complexity to grasp selection and execution \cite{5zhai20222}.

While recent advancements have addressed aspects of this complex problem, a unified solution remains elusive. Traditional grasp synthesis often lacks task-specificity and struggles with explicit constraints \cite{1mousavian20196,2sundermeyer2021contact}. Vision-Language Models (VLMs) and Large Language Models (LLMs) have significantly improved reasoning for task understanding and affordance grounding \cite{10tang2025affordgrasp,11qian2024thinkgrasp}. However, these reasoning-focused methods typically rely on downstream general-purpose grasp generators \cite{12fang2020graspnet,13fang2023anygrasp} that may not be optimized for precise grasping within VLM-identified constrained regions. Conversely, specialized constrained grasp generation techniques, like VCGS \cite{14lundell2023constrained}, often require extensive annotated datasets. More recently, Constrained Grasp Diffusion Fields (CGDF) \cite{15singh2024constrained} introduced a powerful Part-Guided Diffusion strategy for sample-efficient constrained grasping without retraining. Yet, CGDF itself lacks the upstream perception and reasoning capabilities to define these constraint regions from open-vocabulary instructions. Moreover, robust, task-oriented dual-arm grasping guided by language remains a developing area, often relying on geometric sampling rather than deeper semantic understanding \cite{5zhai20222}.

% 使用单栏小图
\begin{figure}[t]
    \centering
    \includegraphics[width=\linewidth]{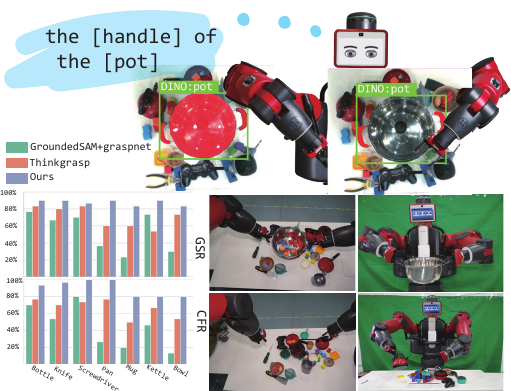}
    \caption{UniDiffGrasp: Open-Vocabulary Constrained Grasping Example and Performance Comparison. UniDiffGrasp innovatively applies a VLM-guided part diffusion strategy, achieving state-of-the-art open-vocabulary part grasping and uniquely extending this precision to complex dual-arm cooperation. Its unified end-to-end framework, integrating single and dual-arm capabilities, demonstrates exceptional real-world performance and strong generalization.}
    \label{fig:1}
\end{figure}
To overcome these limitations, we innovatively propose UniDiffGrasp, a unified framework for open-vocabulary, task-oriented constrained grasping proficient in both single-arm precision and cooperative dual-arm manipulation. UniDiffGrasp uniquely integrates the semantic reasoning power of VLMs with the geometrically precise and efficient grasp generation capabilities of CGDF. Our system first employs a VLM to interpret multimodal user input (language and vision), identifying the target object and its functionally relevant part(s). These semantic targets are then accurately localized using open-vocabulary segmentation techniques. The core innovation lies in directly leveraging the VLM-identified part as a geometric constraint to guide CGDF’s Part-Guided Diffusion strategy. This approach empowers UniDiffGrasp to efficiently generate high-quality, 6-DoF grasp poses precisely within the desired functional region, even for complex geometries, all without requiring constraint-specific retraining. For dual-arm scenarios, UniDiffGrasp extends this approach by defining distinct target regions (guided by VLM reasoning or geometric cues) and applying the Part-Guided Diffusion process for each arm, followed by a rigorous selection module to ensure stable cooperative grasps. Our contributions are as follows:
\begin{itemize}
\item We present an end-to-end framework that seamlessly connects advanced VLM-driven task understanding and affordance foundations with state-of-the-art diffusion-based constrained grasp generation. The framework is capable of interpreting implicit human commands and performing precise grasps in complex environments, targeting specified object portions for both one-arm and coordinated two-arm tasks.
\item We demonstrate the effective application of the Part-Guided Diffusion strategy guided directly by VLM-inferred semantic constraints, enabling zero-shot, task-oriented grasping on targeted regions of novel objects without constraint-specific retraining.
\item We introduce a principled methodology for coordinated dual-arm affordance grasping, incorporating flexible target region definition, independent constrained grasp generation per arm via Part-Guided Diffusion, and robust pair selection for kinematic feasibility and stability.
\item We validate UniDiffGrasp through extensive experiments, showcasing state-of-the-art performance in grasp success, sample efficiency, and robustness, particularly in challenging scenarios involving complex geometries, constraints, and dual-arm coordination on a physical robot platform.
\end{itemize}

\begin{figure*}
    \centering
    \includegraphics[width=\linewidth]{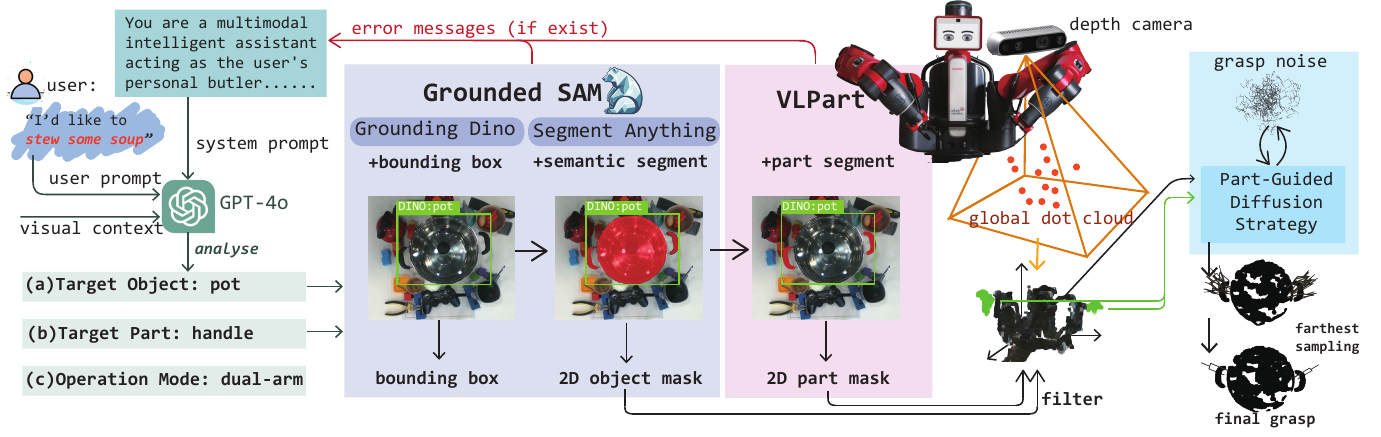}
    \caption{ UniDiffGrasp Framework: Integrating VLM Reasoning, Multi-Stage Segmentation, and Part-Guided Diffusion for Grasp Generation. User input (e.g., `I'd like to stew some soup') and visual context are first processed by a VLM (GPT-4o) to determine the target object (`pot'), functional part (`handle'), and operation mode (`dual-arm'). Multi-stage segmentation—Grounded SAM for object localization and VLPart for precise part identification—then grounds these semantic targets into geometric constraints. Finally, these constraints guide the Part-Guided Diffusion strategy, employing the dual-arm coordination method (Sec. III-C) to define distinct target regions (e.g., for each handle of the pot) and select stable cooperative grasps.}
    \label{fig:2}
\end{figure*}

\section{RELATED WORK}

\subsection{Task-Oriented and Affordance-Based Grasping}
Moving beyond simple object acquisition, task-oriented grasping aims to select grasp poses facilitating subsequent actions by targeting specific functional parts or affordances \cite{1mousavian20196,2sundermeyer2021contact}. Early approaches often relied on predefined object models or known functionalities \cite{18sahbani2012overview}, limiting generalization. Learning-based methods improved this by learning affordances from demonstrations \cite{19bahl2023affordances} or transferring skills via similarity \cite{20ju2024robo}, though often requiring extensive data or complex retrieval.

More recently, Large Language Models (LLMs) and Vision-Language Models (VLMs) have enhanced task context understanding. For instance, AffordGrasp \cite{10tang2025affordgrasp} excels at interpreting implicit user instructions and grounding task-relevant affordances onto object parts using VLMs, demonstrating strong in-context reasoning. Systems like GraspGPT \cite{21tang2023graspgpt} and FoundationGrasp \cite{22tang2025foundationgrasp} also leverage LLMs/VLMs to infer appropriate grasp regions. However, these reasoning-focused methods often output a semantic region that is then passed to a general-purpose grasp detector downstream. This decoupling can lead to inefficiencies or failures if the downstream generator is not optimized for the precise geometric constraints of the inferred affordance, especially for complex shapes. While the concept of ``affordance grounding" \cite{23chen2023affordance} is crucial for localizing interactable regions (e.g., via segmentation masks or heatmaps \cite{25ma2024glover}) and identifying functional parts, it does not inherently guarantee efficient generation of valid 6-DoF grasps within those specific parts. To address these limitations,  UniDiffGrasp integrates advanced VLM-driven affordance identification with a specialized part-guided diffusion strategy for constrained grasp synthesis.

\subsection{Vision-Language Models  for Robotic Grasping and Scene Understanding}
The integration of Vision-Language Models (VLMs) has revolutionized open-vocabulary robotic interaction. VLMs excel at interpreting free-form natural language, inferring user intent even from implicit commands (e.g., ``I'm thirsty" implying a graspable cup handle) \cite{10tang2025affordgrasp,26tong2024oval}. A key capability is grounding these language descriptions to specific objects or regions within a visual scene \cite{27liu2024grounding}. This is crucial for identifying the target object and its relevant parts, often achieved by leveraging open-vocabulary segmentation techniques like SAM \cite{28kirillov2023segment} for general object delineation or part-specific models like VLPart \cite{17sun2023going} for finer-grained component identification.

Several works directly apply VLMs to grasping. For instance, LAN-Grasp \cite{29mirjalili2024lan} uses LLMs to identify graspable parts, OVAL-Prompt \cite{30tong2024oval} frames affordance identification as a VLM grounding task, and ThinkGrasp \cite{11qian2024thinkgrasp} showcases VLM reasoning for strategic clutter removal. While frameworks like AffordGrasp \cite{10tang2025affordgrasp} provide strong VLM-based affordance reasoning, a recurring limitation in many VLM-driven grasping pipelines, as highlighted in Section II-A, remains the decoupling between semantic understanding and low-level grasp generation. The VLM typically outputs a target region , which is then passed to a separate grasp synthesis module (GraspNet \cite{12fang2020graspnet}, AnyGrasp \cite{13fang2023anygrasp}). This downstream module may struggle to efficiently generate poses satisfying the precise VLM-imposed constraints, especially for complex affordance geometries, leading to low sample efficiency or failures. UniDiffGrasp addresses this bottleneck by tightly integrating VLM-derived geometric constraints directly into a specialized constrained grasp generator.

\subsection{Constrained Grasp Generation}
Generating 6-DoF grasp poses within a predefined sub-region is more challenging than unconstrained grasping, where standard generators like Contact-GraspNet \cite{2sundermeyer2021contact} or AnyGrasp \cite{13fang2023anygrasp} find grasps anywhere on the object, often unsuitable for part-specific tasks.
A common approach to constrained grasping is post-filtering: generating a large set of unconstrained grasps and selecting those that happen to fall within the desired region.This is inherently sample-inefficient, particularly when the target region is small relative to the whole object. While learning-based methods like VCGS \cite{14lundell2023constrained} train models specifically for constrained grasping, they require large-scale annotated datasets, limiting scalability and adaptability to arbitrary runtime constraints.
Diffusion models \cite{31urain2023se} have emerged as powerful for grasp synthesis, though adapting them for constrained generation can be complex. A significant advancement is Constrained Grasp Diffusion Fields (CGDF) \cite{15singh2024constrained}. CGDF's Part-Guided Diffusion strategy uniquely enables a model trained solely on unconstrained data to generate grasps constrained to a target region (e.g., a partial point cloud) during inference. This facilitates zero-shot adaptation to arbitrary constraints, high sample efficiency, and proficient handling of complex object geometries, aided by effective shape representation (e.g., using convolutional plane features \cite{32peng2020convolutional}). UniDiffGrasp directly leverages CGDF's Part-Guided Diffusion as its core grasp generation engine, providing the crucial link between VLM-derived semantic constraints and efficient, geometrically accurate grasp synthesis.

\section{METHODS}
We introduce UniDiffGrasp, a unified framework for open-vocabulary, task-oriented constrained grasping, excelling in both single-arm precision and cooperative dual-arm manipulation. UniDiffGrasp synergizes the high-level semantic reasoning of VLMs with the precise geometric grasp generation of CGDF\cite{15singh2024constrained}. This integration leverages a novel part-conditioned diffusion strategy for enhanced sample efficiency, enabling robust grasping of specified object regions, even for large objects requiring dual-arm coordination. Figure 2 provides an overview of the UniDiffGrasp framework. The subsequent sections detail its core components: vision-language grasp reasoning (Section \ref{sec:vlm_reasoning}), single-arm grasp generation (Section \ref{sec:single_arm_gen}), and dual-arm constrained grasping (Section \ref{sec:dual_arm_gen}).

\subsection{Vision-language grasp reasoning and grounding}
\label{sec:vlm_reasoning}
Effectively interpreting high-level, often ambiguous, open-vocabulary user instructions and accurately grounding them to specific object regions is paramount for task-oriented grasping in cluttered environments. UniDiffGrasp achieves this via an integrated reasoning and grounding pipeline.

Multimodal Instruction Interpretation: The process commences with a VLM, such as GPT-4o \cite{33hurst2024gpt}, interpreting the user's natural language instruction (L) within the visual context of an RGB scene image (I). The VLM analyzes this combined input to distill core interaction parameters: the primary Target Object (O), the functionally relevant Target Part ($p*$) (e.g., the 'handle' for 'pouring'), and the required Operation Mode (single-arm or dual-arm), based on task and object properties. This yields a structured output:
\begin{equation} \label{eq:vlm_output}
\{O,\, p^*,\, \mathrm{Mode}\} = \mathrm{VLM}(L,\, I)
.\end{equation}

Hierarchical Visual Grounding: Building on the VLM's output, this stage translates the identified object and part into precise geometric representations (point clouds) essential for grasp generation. This is achieved in two steps for enhanced accuracy, particularly in clutter:

Object Segmentation: An open-vocabulary segmentation model, GroundedSAM \cite{16ren2024grounded}, processes the input image (I) and the VLM-derived target object identifier (O) to generate a pixel-level mask $M_a$ for the entire target object instance. This isolates the object of interest, from which the global object point cloud $P$ is constructed using corresponding depth data.
\begin{equation} \label{eq:groundedsam}
M_a = \mathrm{GroundedSAM}(I,\, O).
\end{equation}

Part Segmentation: To precisely localize the target part $p*$ on the segmented object, a masked view of the scene $M_{Ma}$ (containing only pixels from $M_a$) is created. An open-vocabulary part segmentation model, VLPart \cite{17sun2023going}, then takes $M_{Ma}$ and the target part description $p*$ as input to predict the final pixel-level part mask $M_p^*$ within the object's boundaries. This hierarchical approach (Object $\rightarrow$ Part) prevents VLPart from incorrectly segmenting similar parts on distractor objects. The target region point cloud $P_t$ is then constructed from depth data corresponding to $M_p^*$. Note that $ P_t \subseteq P$.
\begin{equation} \label{eq:vlpart}
M_{\!p}^* = \mathrm{VLPart}(M_{\!\mathrm{Ma}},\, p^*).
\end{equation}

This reasoning and grounding module effectively transforms potentially ambiguous user input into the essential geometric inputs for subsequent grasp generation: the global object point cloud $P$ and the specific target region point cloud $P_t$.

\subsection{Grasp Generation and Estimation}
\label{sec:single_arm_gen}
Once the vision-language reasoning and grounding stage (Section~\ref{sec:vlm_reasoning}) has yielded the global object point cloud $P$ and the target part point cloud $P_t$, UniDiffGrasp generates 6-DoF grasp poses if the determined Mode is ``single-arm". This stage focuses grasp generation onto the specified affordance region $P_t$ while ensuring global validity with respect to $P$. This is achieved efficiently, without constraint-specific retraining, by leveraging the Part-Guided Diffusion strategy from Constrained Grasp Diffusion Fields (CGDF) \cite{15singh2024constrained}, which builds on learning grasp distributions as energy functions on SE(3) \cite{31urain2023se}.

\textbf{CGDF Core Principles.}
CGDF models the distribution of stable, collision-free grasp poses $H \in \text{SE}(3)$ as a smooth energy landscape, where lower energy $e = E_\theta(H, k, P_{\text{obj}})$ signifies a higher quality grasp. Grasp generation is an inverse diffusion process, iteratively refining poses towards lower-energy regions guided by a score function $\mathbf{s}_\theta = -\nabla_H E_\theta$ \cite{31urain2023se}.
The key innovation, Part-Guided Diffusion \cite{15singh2024constrained}, enables an energy model $E_\theta$ trained only on unconstrained data to generate grasps constrained to a target region $P_t \subseteq P_{\text{obj}}$ during inference. At each diffusion step $k$ for a candidate grasp $H_k$:
\begin{itemize}
    \item \textit{Dual Energy Evaluation:} The energy model $E_\theta$ evaluates $H_k$ with respect to both the global object $P$ (yielding $e_k'$) and the local target part $P_t$ (yielding $e_k''$).
    \item \textit{Guided Energy and Score:} The effective guiding energy is $e_{k, \text{guided}} = \max(e_k', e_k'')$ (Eq. 6 in \cite{15singh2024constrained}). The conditional score vector $\mathbf{s}_{\theta, \text{guided}}(H_k, k, P, P_t)$ dynamically adapts, steering the pose based on whether $e_k'$ or $e_k''$ dominates, to satisfy the more critical constraint (Eq. 7 in \cite{15singh2024constrained}).
\end{itemize}
This focuses generation onto $P_t$ while maintaining global validity.

\textbf{UniDiffGrasp Implementation: Part-Guided Generation and Selection.}
UniDiffGrasp directly employs the Part-Guided Diffusion strategy using the global object point cloud $P$ and the target part point cloud $P_t$ (derived in Section~\ref{sec:vlm_reasoning}) as conditioning inputs for the CGDF inference process.

Starting from random initial grasps, UniDiffGrasp iteratively refines them over $T$ diffusion steps. In each step $k$, candidate grasps $H_k^{(i)}$ are updated towards lower-energy configurations by applying the conditional score vector $\mathbf{s}_{\theta, \text{guided}}(H_k^{(i)}, k, P, P_t)$ (derived from the dual energy evaluation on $P$ and $P_t$, as per Eq. 7 in \cite{15singh2024constrained}). This iterative process, guided by both local ($P_t$) and global ($P$) geometric considerations, yields a set of candidate grasp poses, $\{H_{\text{gen}}\} = \{H_0^{(i)}\}$, which are concentrated on or near the target region $P_t$ while being globally collision-free and stable.

From this refined set $\{H_{\text{gen}}\}$, UniDiffGrasp selects the optimal single-arm grasp $H^*$. As Part-Guided Diffusion has already ensured local targeting and global validity, the final selection prioritizes overall grasp stability by choosing the grasp with the minimum energy when evaluated against the global object point cloud $P$ at noise level $k=0$:
\begin{equation} \label{eq:final_grasp_selection_refined}
H^* = \underset{H_i \in \{H_{\text{gen}}\}}{\operatorname{argmin}} E_\theta(H_i, 0, P).
\end{equation}

The resulting $H^*$ is the final, stability-optimized grasp pose, accurately targeted to the VLM-identified functional part.
\begin{figure}
    \centering
    \includegraphics[width=\linewidth]{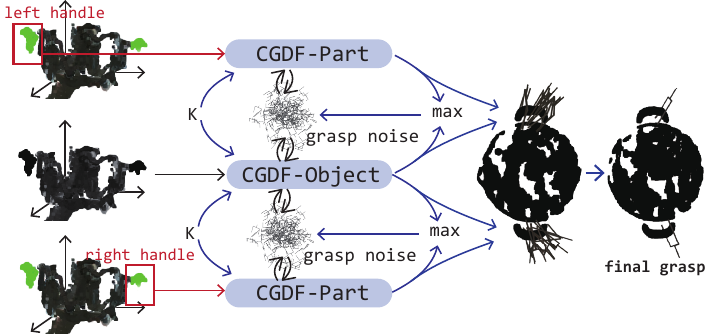}
    \caption{Principle of Dual-Arm Grasp Generation in UniDiffGrasp using Coordinated Part-Guided Diffusion. We illustrate how UniDiffGrasp generates dual-arm grasps, taking a pot with two handles as an example.}
    \label{fig:3}
\end{figure}

\subsection{Constraint Grasping for Large Objects: Dual-Arm Coordination}
\label{sec:dual_arm_gen}
When the VLM-determined Mode (Section~\ref{sec:vlm_reasoning}) is ``dual-arm", typically for large or unwieldy objects, UniDiffGrasp employs its dual-arm coordination strategy, the principle of which is depicted in Fig.~\ref{fig:3}. This extends constrained grasping to enable stable handling by coordinating two robotic arms, beginning with defining two distinct target regions, $P_{t1}$ and $P_{t2}$.

\textbf{Dual Target Region Definition.}
UniDiffGrasp defines $P_{t1}$ and $P_{t2}$ using a context-aware, VLM-guided approach, rather than relying solely on geometric sampling:
\begin{itemize}
    \item \textit{Semantic Splitting:} If the VLM's interpretation of the instruction $L$ and identified target part $p^*$ suggests functionally distinct interaction points for two arms (e.g., $p^*$ identifies ``handles" for a pot), UniDiffGrasp uses the part segmentation model (like VLPart) to locate these semantic parts ($p^*_1, p^*_2$, derived from $p^*$) on the global object point cloud $P$. The resulting segmented point clouds form $P_{t1}$ and $P_{t2}$:
    \begin{equation} \label{eq:semantic_split}
    (P_{\!t1},\, P_{\!t2}) = \mathrm{SegSemParts}(P,\, p_1^*,\, p_2^*).
    \end{equation}
    \item \textit{Geometric Splitting:} If the VLM provides no strong cue for distinct functional parts (e.g., $p^*$ refers to a whole ``keyboard"), or if only a single large part is identified, UniDiffGrasp defaults to a geometric division. The global object point cloud $P$ (or the identified large part $P_t$) is divided into two non-overlapping or minimally overlapping sub-regions $P_{t1}$ and $P_{t2}$ (left/right halves):
    \begin{equation} \label{eq:geometric_split}
    (P_{\!t1},\, P_{\!t2}) = \mathrm{GeometricSplit}(P \text{ or } P_t).
    \end{equation}
\end{itemize}

This flexible process yields the two target regions, $P_{t1}$ and $P_{t2}$. These are determined using the function $\operatorname{DetermineTargetRegions}$ with inputs $P, O, p^*,$ and the VLM Output. These regions then serve as specific geometric constraints for each arm.

\textbf{Independent Constrained Grasp Generation.}
UniDiffGrasp then leverages the Part-Guided Diffusion strategy (Section~\ref{sec:single_arm_gen}) independently for each arm. Arm 1 generates a set of $N$ candidate grasps $\{H1^{(i)}\}$ conditioned on $P$ and its specific target $P_{t1}$. Arm 2 similarly generates $N$ candidates $\{H2^{(j)}\}$ using $P$ and $P_{t2}$. This parallel process produces $N \times N$ potential grasp pairs $(H1^{(i)}, H2^{(j)})$.

\textbf{Optimal Dual-Arm Grasp Pair Selection.}
Selecting the best pair $(H1^*, H2^*)$ from the $N \times N$ potential grasp pairs involves a sequential filtering and selection process:

\begin{enumerate}
    \item \textit{Individual Grasp Filtering:} Candidate grasps for each arm, $\{H1\}$ and $\{H2\}$, are initially filtered. A grasp $H1^{(i)}$ is retained in $H1_{\text{filt}}$ if its energy $E_\theta(H1^{(i)}, 0, P) < \delta$ and $E_\theta(H1^{(i)}, 0, P_{t1}) < \delta$. $H2_{\text{filt}}$ is obtained similarly, using $P_{t2}$. $\delta$ is a predefined energy threshold.
\begin{flalign} \label{eq:filter_h1_h2} % 标签可以更新，例如_no_fc可以去掉或修改
&H1_{\mathrm{filt}} = \Bigl\{ H1^{(i)} \in \{H1\} \,\Big| \nonumber \\
&\quad E_\theta(H1^{(i)},0,P)<\delta \land E_\theta(H1^{(i)},0,P_{t1})<\delta \Bigr\}, & \\
&H2_{\mathrm{filt}} = \Bigl\{ H2^{(j)} \in \{H2\} \,\Big| \nonumber \\
&\quad E_\theta(H2^{(j)},0,P)<\delta \land E_\theta(H2^{(j)},0,P_{t2})<\delta \Bigr\}. &
\end{flalign}

    \item \textit{Inter-Gripper Collision Checking:} All pairs $(H1_i, H2_j)$ formed from $H1_{\text{filt}}$ and $H2_{\text{filt}}$ are checked for inter-gripper collisions. Non-colliding pairs constitute the set $\text{Pairs}_{\text{no-collision}}$.

     \item \textit{Force Closure (FC) Stability Check:} Each pair $(H1_i, H2_j)$ from $\text{Pairs}_{\text{no-collision}}$ is then rigorously evaluated for grasp stability using a force closure (FC) metric. This assessment considers the combined ability of the two grippers to resist arbitrary external wrenches (forces and torques) applied to the object. We utilize a quantitative FC quality score to assess stability, where pairs falling below a certain threshold are deemed unstable and discarded [\cite{34liu2021synthesizing}]. The remaining pairs form the set $\text{Pairs}_{\text{stable}}$.

    \item \textit{Final Selection via Maximal Distance:} From the set $\text{Pairs}_{\text{stable}}$ ( pairs that are both non-colliding and force-closure stable), the pair $(H1^*, H2^*)$ that maximizes the Euclidean distance $D_{ij}$ between gripper centers (projected onto the object surface) is chosen as the final dual-arm grasp. This promotes balance and stable handling, where $D_{ij} = \text{Dist}(\text{Center}(H1_i), \text{Center}(H2_j))$. The selection is:
    \begin{equation} \label{eq:max_dist_select_fc} % 标签更新
    (H1^*, H2^*) = \operatorname*{argmax}_{(H1_i, H2_j) \in \text{Pairs}_{\text{stable}}} D_{ij}.
    \end{equation}
\end{enumerate}

This integrated approach, combining VLM-guided region definition with independent constrained grasp generation and a stringent selection prioritizing stability and spatial balance, allows UniDiffGrasp to effectively address coordinated dual-arm manipulation.

\section{EXPERIMENTS}
We evaluated the practical efficacy and robustness of UniDiffGrasp through real-world experiments. Our setup included a dual-arm Baxter robot with parallel-jaw grippers and an Intel RealSense D435 camera for scene observation (1280×720 RGB-D). The UniDiffGrasp system, encompassing reasoning and generation modules, is deployed on a server equipped with dual NVIDIA RTX 3090 GPUs.
To quantitatively assess performance, we employ two key metrics:
\begin{itemize}
    \item Grasp Success Rate (GSR): The percentage of successful physical grasp executions (lift and stable hold) out of total attempts,evaluating end-to-end system performance.
    \item Collision-Free Rate (CFR): The percentage of generated grasp poses that result in no collision between the gripper model and the target object mesh. This assesses the geometric feasibility and precision of the grasp generation component itself, especially its ability to respect object geometry within a constrained target region.
\end{itemize}

\subsection{Part-Guided Constrained Grasping Experiments}
This section empirically evaluates UniDiffGrasp's performance in single-arm tasks requiring grasps constrained to specific functional object parts, comparing its Part-Guided Diffusion strategy against relevant baselines.

UniDiffGrasp, as detailed in Section IV-B and IV-A, first uses VLM+segmentation to acquire global (P) and target part (Pt) point clouds. It then employs the Part-Guided Diffusion strategy to generate grasps focused on Pt while respecting P, selecting the final grasp via minimal global energy.

We compare UniDiffGrasp against two baselines:
\begin{itemize}
    \item \textbf{GroundedSAM + GraspNet}: This baseline uses GroundedSAM \cite{16ren2024grounded} for global object segmentation (P) and GraspNet \cite{13fang2023anygrasp} for unconstrained grasp generation on P.
    \item \textbf{ThinkGrasp \cite{11qian2024thinkgrasp}}: A SOTA baseline that utilizes VLPart \cite{17sun2023going} for part segmentation and GraspNet \cite{13fang2023anygrasp} for grasp generation, with grasp selection performed as described in their original work.
\end{itemize}
Experiments involved 30 trials per method on seven diverse household objects using a Baxter robot. Each trial included: (1) generating candidate grasps, (2) selecting the top-ranked grasp, (3) evaluating CFR against the ground-truth mesh, (4) executing the grasp, and (5) recording GSR.

The results (Table 1) unequivocally demonstrate UniDiffGrasp's superior performance.
\textbf{UniDiffGrasp achieves an outstanding average GSR of 0.876 and a CFR of 0.900.} This performance significantly surpasses the current state-of-the-art baseline, ThinkGrasp, and markedly exceeds the simpler GroundedSAM+GraspNet approach. This firmly establishes UniDiffGrasp as a new state-of-the-art method for constrained, task-oriented grasping, showcasing substantially higher success rates and geometric accuracy, especially for complex objects like Pans, Mugs, and Kettles where baselines falter. The GroundedSAM+GraspNet struggles due to its lack of part-specific reasoning. While ThinkGrasp incorporates part segmentation, its reliance on a general-purpose downstream grasp generator (GraspNet) still limits its precision within the constrained region. In contrast, UniDiffGrasp's tight integration of precise VLM-driven part identification with the specialized Part-Guided Diffusion of CGDF ensures grasps are both accurately targeted and globally valid, leading to higher GSR and CFR. Visualizations are in Fig. 4.

% Table 1: 单臂抓取实验结果

\begin{table*}[t]
\centering
\caption{Single-Arm Grasping Performance Comparison}
\label{tab:single}
\small % 使用小字号
\resizebox{\textwidth}{!}{
\begin{tabular}{@{} l *{7}{S[table-format=1.3] S[table-format=1.3]} S[table-format=1.3] S[table-format=1.3] @{}}
\toprule
\multirow{2}{*}{Method} & 
\multicolumn{2}{c}{Bottle} & \multicolumn{2}{c}{Knife} & 
\multicolumn{2}{c}{Screwdriver} & \multicolumn{2}{c}{Pan} & 
\multicolumn{2}{c}{Mug} & \multicolumn{2}{c}{Pliers} & 
\multicolumn{2}{c}{Kettle} & 
\multicolumn{2}{c}{Average} \\
\cmidrule(lr){2-3} \cmidrule(lr){4-5} \cmidrule(lr){6-7} 
\cmidrule(lr){8-9} \cmidrule(lr){10-11} \cmidrule(lr){12-13} 
\cmidrule(lr){14-15} \cmidrule(lr){16-17} 
& {GSR} & {CFR} & {GSR} & {CFR} & {GSR} & {CFR} & {GSR} & {CFR} & {GSR} & {CFR} & {GSR} & {CFR} & {GSR} & {CFR} & {GSR} & {CFR} \\
\midrule
GSG & 0.767 & 0.700 & 0.667 & 0.533 & 0.700 & 0.800 & 0.367 & 0.267 & 0.233 & 0.200 & 0.733 & 0.467 & 0.300 & 0.133 & 0.538 & 0.443 \\
TGrasp & 0.833 & 0.767 & 0.800 & 0.700 & 0.833 & 0.733 & 0.600 & 0.767 & 0.600 & 0.500 & 0.533 & 0.667 & 0.733 & 0.533 & 0.705 & 0.667 \\
Ours & \textbf{0.900} & \textbf{0.933} & \textbf{0.900} & \textbf{0.967} & \textbf{0.867} & \textbf{1.000} & \textbf{0.900} & \textbf{1.000} & \textbf{0.833} & \textbf{0.800} & \textbf{0.900} & \textbf{0.800} & \textbf{0.833} & \textbf{0.800} & \textbf{0.876} & \textbf{0.900} \\
\bottomrule
\end{tabular}
}
\end{table*}

\begin{figure*}[!t]
    \centering
    \includegraphics[width=\linewidth]{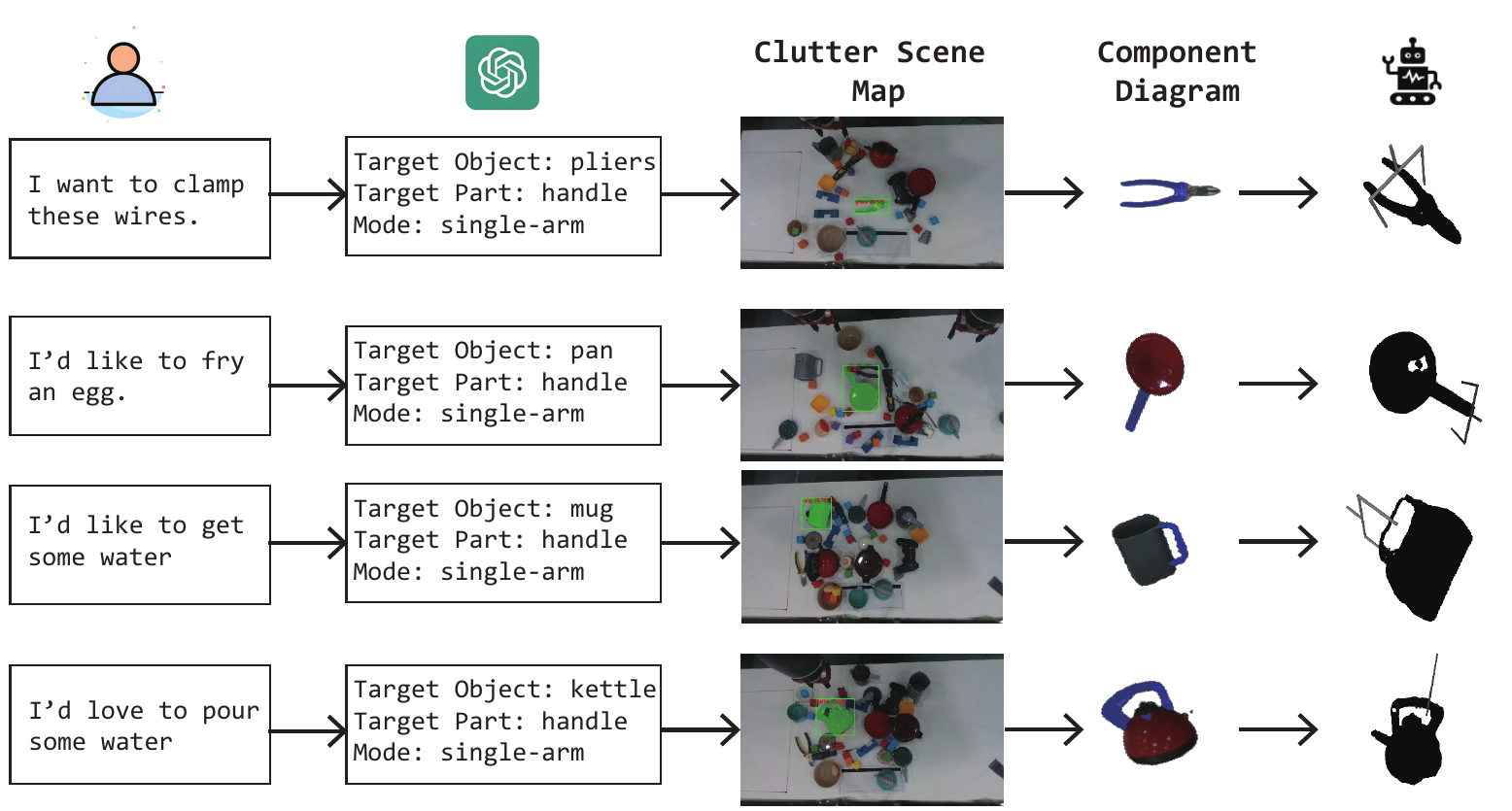}
    \caption{Visualization of task-oriented grasping of functional parts by single-arm}
    \label{fig:4}
\end{figure*}
%\vspace{-1.5em} % 减少第一张图与第二张图之间的间距

\subsection{Dual-Arm Large Object Grasping Experiments}
\label{sec:dual_arm_exp}

This section evaluates UniDiffGrasp's capability for cooperative dual-arm manipulation of large-scale objects, highlighting the benefits of its integrated approach.

The methodology for UniDiffGrasp in dual-arm scenarios, as detailed in Section~\ref{sec:dual_arm_gen} (Dual-Arm Coordination), involves three main steps. First, \textbf{Target Region Definition} is performed either via \textit{Semantic Splitting} (using VLPart to segment functional parts like handles if VLM cues are clear) or \textit{Geometric Splitting} (dividing the global point cloud $P$ into left/right halves if no distinct semantic parts are identified). Second, \textbf{Independent Grasp Generation} for each arm utilizes Part-Guided Diffusion conditioned on their respective target regions ($Pt_1, Pt_2$). Finally, \textbf{Optimal Pair Selection} is achieved through a multi-stage process including energy threshold filtering , inter-gripper collision checking, maximal center distance prioritization, and force closure validation.

For comparison, the baseline method employs a strategy similar to that described in the original CGDF paper for dual-arm grasping: it uses Farthest Point Sampling (FPS) followed by K-Nearest Neighbors (KNN) to randomly generate two target regions from the global object point cloud, upon which grasp generation is performed. This baseline lacks semantic guidance for region definition and advanced pair selection heuristics.

The experimental setup was consistent with Section A. Four categories of large items were tested, with 30 execution trials performed per object-method combination.

% 双臂抓取性能表 (Table 2)
\begin{table*}[t]
\centering
\caption{Comparison OF Dual-Arm Grasping Performance }
\label{tab:dual}
\small % 使用小字号
\begin{tabular}{@{} l *{4}{S[table-format=1.3] S[table-format=1.3]} S[table-format=1.3] S[table-format=1.3] @{}}
\toprule
\multirow{2}{*}{Method} & 
\multicolumn{2}{c}{Basin} & \multicolumn{2}{c}{Keyboard} & 
\multicolumn{2}{c}{Laptop} & \multicolumn{2}{c}{Pot} & 
\multicolumn{2}{c}{Average} \\
\cmidrule(lr){2-3} \cmidrule(lr){4-5} \cmidrule(lr){6-7} \cmidrule(lr){8-9} \cmidrule(lr){10-11}
& {GSR} & {CFR} & {GSR} & {CFR} & {GSR} & {CFR} & {GSR} & {CFR} & {GSR} & {CFR} \\
\midrule
Baseline & 0.633 & 0.300 & 0.700 & 0.333 & 0.367 & 0.300 & 0.200 & 0.133 & 0.475 & 0.267 \\
Ours & \textbf{0.800} & \textbf{0.867} & \textbf{0.767 }& \textbf{0.833} & \textbf{0.767} & \textbf{0.800} & \textbf{0.733} & \textbf{0.900} & \textbf{0.767} & \textbf{0.850} \\
\bottomrule
\end{tabular}
\end{table*}

\begin{figure*}
    \centering
    \includegraphics[width=\linewidth]{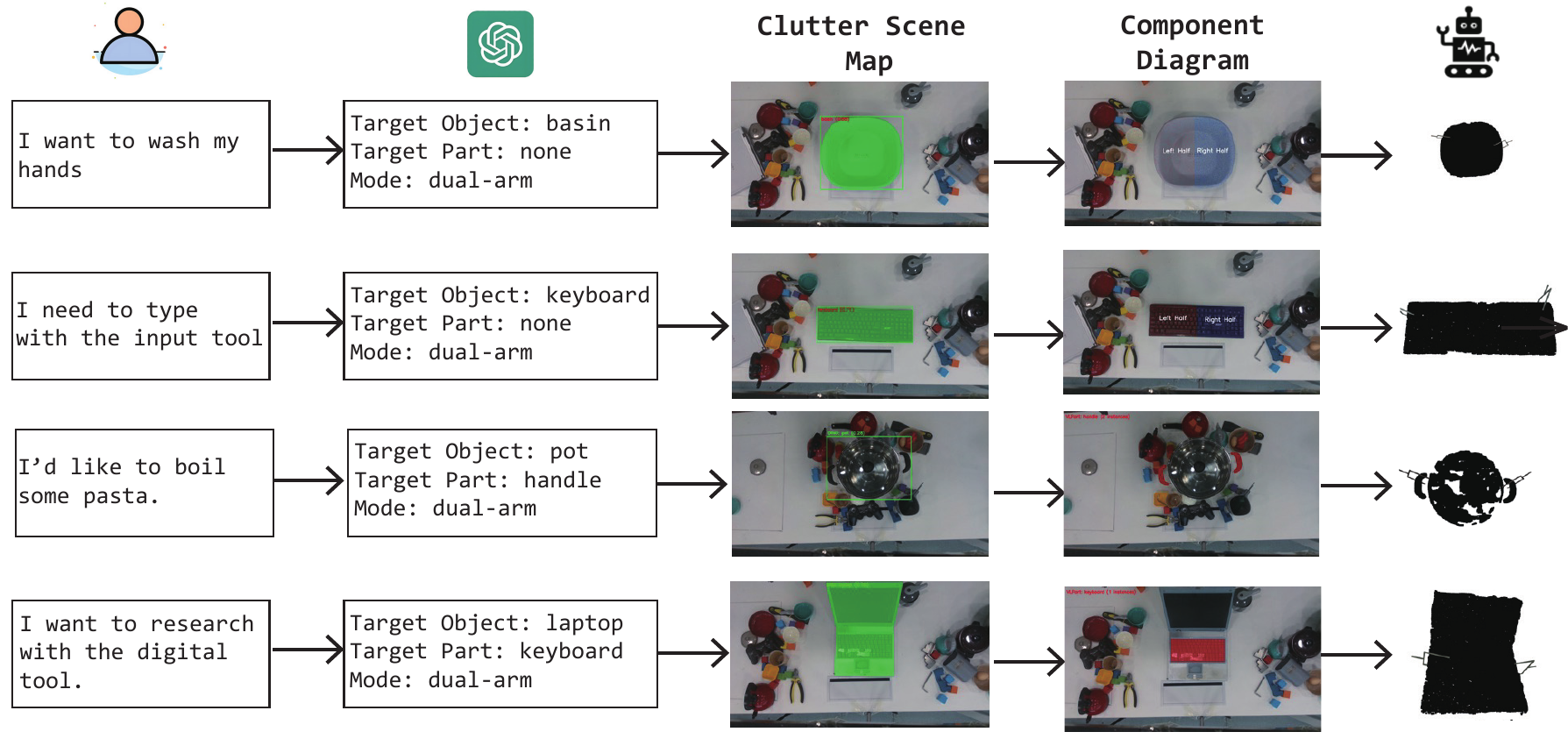}
    \caption{Visualization of collaborative dual-arm grasping for large object manipulation}
    \label{fig:5}
\end{figure*}

The quantitative results, summarized in Table~\ref{tab:dual}, clearly demonstrate the advantages of UniDiffGrasp.

UniDiffGrasp achieves a significantly higher average GSR of 0.767 and a vastly superior average CFR of 0.850 across all tested objects.
For complex objects like the Pot and Laptop, which often possess specific functional parts or non-trivial geometries for dual-arm handling, UniDiffGrasp's ability to leverage semantic understanding (for the Pot's handles) or apply robust geometric splitting combined with sophisticated pair selection yields markedly superior performance,whereas the baseline struggles significantly.

Even for simpler, more symmetric objects like the Basin and Keyboard, where semantic cues for distinct parts might be less critical, UniDiffGrasp still demonstrates substantially better GSR and CFR. This indicates that UniDiffGrasp's structured approach to geometric region definition and its advanced optimal pair selection (including maximal distance and force closure) contribute to more stable and feasible grasps compared to the baseline's random region generation and simpler selection.Visualizations are in Fig. 5.

\section{CONCLUSIONS}

This paper introduces UniDiffGrasp, a novel framework significantly advancing open-vocabulary, task-oriented constrained grasping for both single and dual-arm manipulation. UniDiffGrasp uniquely integrates VLM semantic reasoning with CGDF's part-guided diffusion, effectively translating high-level instructions into accurate grasps on specific functional parts. This is achieved by leveraging VLM-identified parts as geometric constraints for efficient, zero-shot grasp generation and employing a principled methodology for stable dual-arm grasping.
Comprehensive real-world experiments validate UniDiffGrasp, demonstrating substantial improvements over state-of-the-art methods. It achieved grasp success rates of 0.876 (single-arm) and 0.767 (dual-arm) with enhanced sample efficiency, particularly in challenging scenarios involving complex geometries and multi-arm coordination. These results highlight UniDiffGrasp's capability for precise, coordinated open-vocabulary grasping in complex real-world settings, underscoring the efficacy of its core integration.

%%%%%%%%%%%%%%%%%%%%%%%%%%%%%%%%%%%%%%%%%%%%%%%%%%%%%%%%%%%%%%%%%%%%%%%%%%%%%%%%

\end{document}